\documentclass{article}
\pdfoutput=1

\usepackage[utf8]{inputenc}
\usepackage[T1]{fontenc}
\usepackage{tgpagella}
\usepackage{mathpazo}
\usepackage{inconsolata}
\usepackage[a4paper,margin=2.5cm,headheight=24pt]{geometry}
\usepackage{booktabs}
\usepackage{amsfonts}
\usepackage{amsmath}
\usepackage{nicefrac}
\usepackage{microtype}
\usepackage{graphicx}
\graphicspath{{media/}}
\usepackage{natbib}
\usepackage{longtable}
\usepackage{array}
\usepackage{calc}
\usepackage{tabularx}
\usepackage{float}
\usepackage{enumitem}
\usepackage{caption}
\usepackage{fancyhdr}
\usepackage{titlesec}
\usepackage{titling}
\usepackage[table]{xcolor}
\usepackage{url}
\usepackage{doi}
\usepackage{hyperref}

\definecolor{darkblue}{rgb}{0,0,0.5}
\definecolor{resultgroup}{HTML}{E8E8E8}
\definecolor{skillrow}{HTML}{EEEEFF}
\definecolor{positive}{HTML}{16894A}
\definecolor{negative}{HTML}{C62828}

\setcitestyle{authoryear,round,citesep={;},aysep={,},yysep={;}}

\makeatletter
\renewcommand{\normalsize}{\@setfontsize\normalsize{11pt}{13pt}}
\renewcommand{\small}{\@setfontsize\small{10pt}{11.5pt}}
\renewcommand{\footnotesize}{\@setfontsize\footnotesize{9.5pt}{11pt}}
\makeatother
\normalsize

\titleformat{\section}{\large\bfseries\raggedright}{\thesection}{1em}{}
\titlespacing*{\section}{0pt}{2.6ex plus 0.5ex minus 0.2ex}{1.8ex plus 0.3ex minus 0.2ex}
\titleformat{\subsection}{\normalsize\bfseries\raggedright}{\thesubsection}{1em}{}
\titlespacing*{\subsection}{0pt}{2.2ex plus 0.5ex minus 0.2ex}{1.1ex plus 0.2ex}
\titleformat{\subsubsection}{\normalsize\bfseries\raggedright}{\thesubsubsection}{1em}{}
\titlespacing*{\subsubsection}{0pt}{2ex plus 0.5ex minus 0.2ex}{0.8ex plus 0.2ex}

\fancypagestyle{plain}{%
  \fancyhf{}%
  \fancyfoot[C]{\thepage}%
}
\fancypagestyle{firstpage}{%
  \fancyhf{}%
  \fancyfoot[C]{\thepage}%
}
\renewcommand{\footnoterule}{\kern-3pt\hrule width 12pc\kern 2.6pt}

\pretitle{{\color[HTML]{4D4D4D}\noindent\rule{\textwidth}{1pt}}\par\vspace{0.25in}\centering\Large\bfseries}
\posttitle{\par\vspace{0.1cm}}
\preauthor{\centering\normalsize}
\postauthor{\par\vspace{0.2cm}}
\predate{}
\postdate{}

\title{AGL-SKILL: Coding Harnesses as Agentic Optimizers}
\author{
  \begin{tabular}{@{}c@{\hspace{4em}}c@{\hspace{4em}}c@{}}
    \textbf{Ryan Tsui\thanks{Work done during an internship at Microsoft Research Asia, Shanghai.}}
      & \textbf{Yuge Zhang}
      & \textbf{Yuqing Yang}
  \end{tabular}\\[0.55em]
  Microsoft Research Asia\\[0.45em]
  \begin{tabular}{@{}c@{\hspace{2em}}c@{\hspace{2em}}c@{}}
    \href{mailto:s2480214@ed.ac.uk}{\texttt{s2480214@ed.ac.uk}}
      & \href{mailto:Yuge.Zhang@microsoft.com}{\texttt{Yuge.Zhang@microsoft.com}}
      & \href{mailto:Yuqing.Yang@microsoft.com}{\texttt{Yuqing.Yang@microsoft.com}}
  \end{tabular}
}
\date{}

\hypersetup{
  pdftitle={AGL-SKILL: Coding Harnesses as Agentic Optimizers},
  pdfauthor={Ryan Tsui, Yuge Zhang, Yuqing Yang},
  pdfkeywords={agent skills, coding agents, agent optimization, empirical evaluation},
  colorlinks=true,
  hypertexnames=false,
  linkcolor=darkblue,
  citecolor=darkblue,
  urlcolor=darkblue
}

\renewenvironment{abstract}
  {\vskip.075in\centerline{\large\bfseries Abstract}\vspace{0.5ex}\begin{quote}}
  {\par\end{quote}\vskip 1ex}

\begin{document}
\maketitle
\thispagestyle{firstpage}

\begin{abstract}

Frontier models are increasingly used through harnesses as agents, with growing autonomy to do useful work. 
Agent performance can often be enhanced by the use of skills: reusable workflows, code, and prompts that provide task-related information or guide the behaviour of the model. 
Skills, being textual in nature, present themselves as another editable surface to be optimized; current methods have improved on handcrafting and one-shot generation by treating skill development as an iterative process, using score-driven search to tweak and improve performance. 
In this paper, we demonstrate the efficacy of the agents themselves as optimizers: across five benchmarks, a coding agent without guidance substantially improves every target and outperforms published skill-optimization methods. 
More broadly, this shows the capability of agents to do such work and to refine their own behaviour to provide task-specific context, as long as they are given a metric to measure against.

\end{abstract}

\section{Introduction}\label{introduction}

\begin{figure}[t]
\centering
\includegraphics[width=\linewidth]{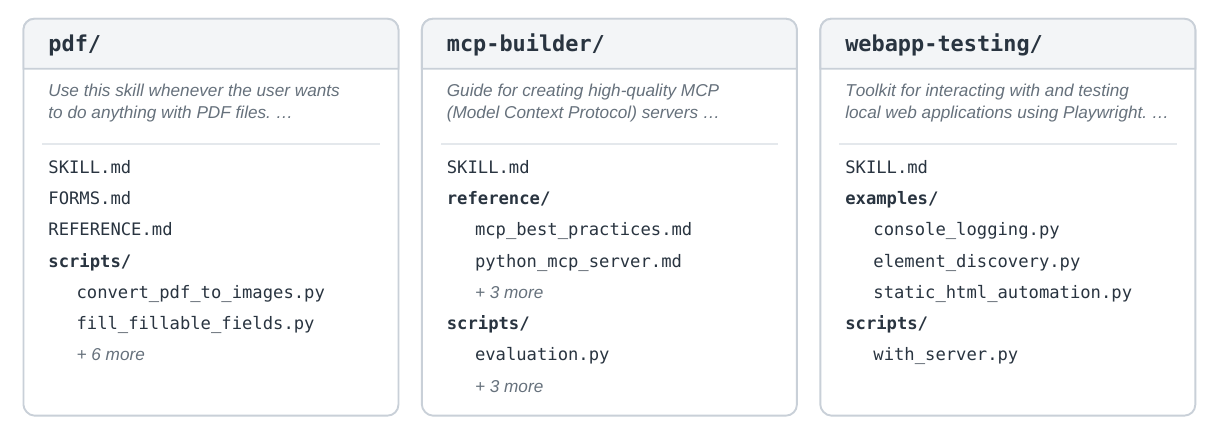}
\caption{Examples of skills from the Anthropic skills repository \citep{anthropicskills}.}
\label{fig:skill-examples}
\end{figure}

Large Language Models (LLMs) are being applied to an increasingly diverse set of tasks, often through harnesses such as Claude Code or Codex to more effectively engage with the engineering environment. Harnesses provide models with tooling, context management, an execution environment, and subagent orchestration for them to better solve the problems they are working on; the combination of the two is commonly referred to as an agent and granted a large degree of autonomy in practice \citep{anthropic2026autonomy}. Additional fine-grained context for individual users and tasks can be provided to agents on demand through skills (Figure~\ref{fig:skill-examples}), which are folders of instructions, scripts, and reference material \citep{agentskills}. How much a skill helps depends on these contents: whether by providing more nuanced information or more refined tooling built off an understanding of the problem space, some skills are better than others.

The challenge, therefore, is to find a way to reliably develop skills that raise agent performance, whatever the task at hand. Several ways have been explored: the most direct is to handcraft the skill, but this is limited by its author and inflexible to a shifting task environment; more recent score-driven methods treat skill development as an optimization problem \citep{yang2026skillopt,alzubi2026evoskill}, using models to tweak the skill in a hill-climbing process that keeps only the edits which improve the score, and is inefficient as a result.

\begin{figure}[t]
\centering
\includegraphics[width=\linewidth]{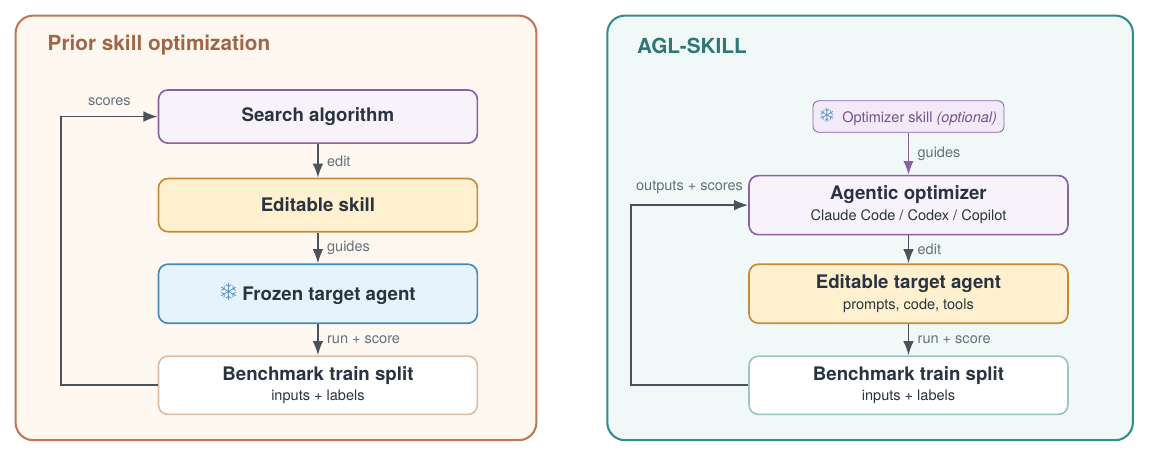}
\caption{A comparison of previous techniques and AGL-SKILL. Snowflakes denote fixed components.}
\label{fig:method-comparison}
\end{figure}

Agents, by contrast, have a wider but directed search space, and already possess the tooling and environment to do this work. We therefore take the skill of a \emph{target} agent as the object to be optimized, and set an \emph{optimizer} agent to iterate on it, free to edit the rest of the target as well, guided by the target's performance on the train split of a benchmark (Figure~\ref{fig:method-comparison}). We call this setup AGL-SKILL. To this end, we evaluate frontier models as agentic optimizers: two state-of-the-art models in three popular harnesses, at three budgets and on five benchmarks. Even without guidance, the agent is already a strong optimizer, outperforming every published method we compare against.

Subsequently, we ask whether guidance improves on this by giving the optimizer the Agent Lightning skill -- which is written for the optimization task itself -- along with variants that use two-phase optimization, subagents, or Git worktrees. The answer is largely no: the skill and the variants occasionally raise the score, but none consistently beats the unguided optimizer by a notable margin. The skill is instead designed to set up the environment for a fresh agent with no prior context of the task. We also report the cost of each optimization and the test-time cost of the resulting target agent. A larger budget does not reliably buy a higher score, and the final score varies from run to run without tracking the optimization cost.

The optimizer traces help explain why guidance adds so little: with or without it, the optimizers arrive at much the same solutions. These solutions are seldom far-fetched; making the target more robust is often enough to score highly, without recourse to an esoteric approach. The Agent Lightning skill is published in the Agent Lightning repository (\url{https://github.com/microsoft/agent-lightning}).

\section{Related work}\label{related-work}

\textbf{Prompt and workflow optimization.} Automatic prompt optimization treats instructions as parameters to be improved from task feedback, by generating and selecting candidates \citep{zhou2023ape}, conditioning on previous candidates and their scores \citep{yang2024opro}, searching over natural-language critiques \citep{pryzant2023protegi}, propagating textual feedback backward \citep{yuksekgonul2024textgrad}, or evolving prompts through reflection \citep{agrawal2025gepa}. DSPy and MIPRO extend this to multi-stage programs \citep{khattab2023dspy,opsahlong2024mipro}, while AgentOptimizer, ADAS, AFlow, and Cognify optimize an agent's functions, code, or workflow structure \citep{zhang2024agentoptimizer,hu2024adas,zhang2024aflow,he2025cognify}. The latter reach beyond prompt text, but each supplies a purpose-built search procedure over a particular agent representation.

\textbf{Skill optimization.} Skill optimization targets reusable agent guidance rather than a single task prompt. SkillOpt trains a single skill document with bounded, validation-gated edits \citep{yang2026skillopt}, EvoSkill uses proposer and builder agents to evolve skills, including helper scripts, for a coding agent \citep{alzubi2026evoskill}, and Trace2Skill distills trajectory-local lessons into transferable guidance \citep{ni2026trace2skill}. These are the closest published baselines for our evaluation. Our optimizer has a wider editable surface: beyond the skill, scripts, and system prompt that EvoSkill can edit, it may also change the target's tools and runtime policy.

\textbf{Iterative and self-improving agents.} Agents have also been made to improve themselves: Self-Refine and Reflexion revise their own outputs from self-generated feedback \citep{madaan2023selfrefine,shinn2023reflexion}, while STOP, SICA, and the Darwin G\"odel Machine edit their own scaffolding or implementation \citep{zelikman2023stop,robeyns2025sica,zhang2025dgm}. Agent Lightning instead separates agent execution from reinforcement-learning training \citep{luo2025agentlightning}. In AGL-SKILL, the optimizer is an off-the-shelf coding agent and the target a separate agent.

\textbf{Coding agents as optimizers.} Most recently, coding agents have themselves been used to improve a separate agent. VeRO wraps a coding agent in a harness of versioned snapshots, budgeted evaluation, and execution traces \citep{ursekar2026vero}; HarnessOpt-Bench measures how well frontier models edit a seed harness under a fixed evaluation budget \citep{ursekar2026harnessoptbench}; CASD has a coding agent distill a static corpus of trajectories into a prompt in a single pass \citep{singh2026casd}; and SkillOpt-Lite reduces skill optimization to a minimal pipeline that extends to the whole harness \citep{shen2026skilloptlite}. Outside coding agents, OEO has a frontier model compose its own skill-improvement loop, which matches or beats SkillOpt and GEPA with a frontier optimizer but not with a medium one \citep{xue2026oeo}. VeRO also varies what the optimizer is given: handing Claude Code its evaluation tools and pattern library moves the average score by 0.02, and the larger gain it reports comes from a different agent built on that library. AGL-SKILL puts the same question to off-the-shelf coding agents, given a skill or a prescribed workflow of their own, and finds that neither consistently helps.

\begin{figure}[H]
\centering
\includegraphics[width=\linewidth]{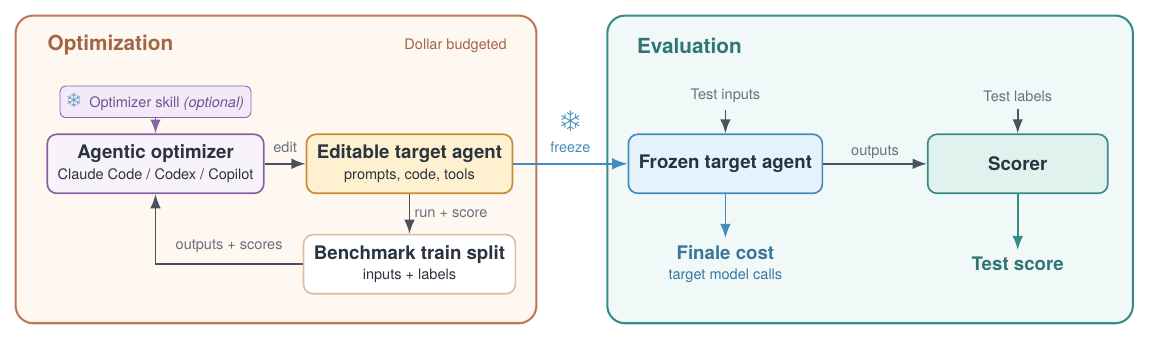}
\caption{Experimental setup of AGL-SKILL: the optimizer iterates on the target agent under a fixed dollar budget, and the frozen target agent is then evaluated on the held-out split.}
\label{fig:evaluation-infrastructure}
\end{figure}

\section{Method}\label{method}

\textbf{Overview.} For each run, we give the optimizer the train split (Figure~\ref{fig:evaluation-infrastructure}). The optimizer accesses the model API through a proxy, which monitors token usage and prevents web access. While within budget, the optimizer can inspect the training examples, edit the target agent's prompts, code, tools, and skills, and implement any other changes. It decides when to run evaluation on the target and which subset of the split to run it on. Once optimization ends -- whether the optimizer chooses to stop or the budget is used up -- the edited target agent is frozen and run on the test split of the benchmark, where a separate evaluator scores the results. We run up to three repetitions of each condition, under the setup summarized in Table~\ref{tab:setup}. The optimizer is free to change the target's effort.

\begin{table}[H]
\centering
\small
\begin{tabular}{@{}ll@{}}
\toprule
Optimizer models & \texttt{claude-opus-5} (Claude Code), \texttt{gpt-5.6-sol} (Codex, GitHub Copilot) \\
\midrule
Target model & \texttt{gpt-5.4-mini} \\
\midrule
Harnesses & Claude Code CLI 2.1.175, OpenAI Codex SDK 0.144.1, GitHub Copilot CLI 1.0.71 \\
\midrule
Budgets & \$5, \$10, \$25 \\
\midrule
Reasoning effort & high (optimizer), medium (target) by default \\
\bottomrule
\end{tabular}
\caption{Experimental setup.}
\label{tab:setup}
\end{table}

\textbf{Optimization variants.} We evaluate five optimizer conditions:

\emph{Vanilla.} The optimizer receives only the task prompt (Appendix~\ref{appendix-prompts}). Every other variant is built on top of this base version.

\emph{Two-phase.} The optimizer first maximizes accuracy, then reduces the target's inference cost while maintaining the score it has reached.

\emph{Agent Lightning skill.} The optimizer loads the published Agent Lightning optimizer skill, which covers budget tracking, the action space of possible changes with tradeoffs, how to read failure evidence, and the evaluation context.

\emph{Subagents.} As an extension of the Agent Lightning skill, the optimizer is told to spread part of its budget across two subagents to explore the solution space in parallel, then to select or combine the best-performing solutions.

\emph{Git worktrees.} On top of the Agent Lightning skill, the optimizer is told to build candidate solutions in separate Git worktrees from the same starting commit, compare them on the same training cases, and restore the best one.

Subagents and Git worktrees are run at the \$10 budget only.

\textbf{Benchmarks.} We evaluate on five benchmarks: SpreadsheetBench \citep{ma2024spreadsheetbench}, ALFWorld \citep{shridhar2021alfworld}, OfficeQA \citep{databricks2025officeqa}, SearchQA \citep{dunn2017searchqa}, and LiveMath \citep{he2026livemathematicianbench}. Table~\ref{tab:evaluation-targets} lists the tasks, splits, and scoring methods.

\begin{table}[H]
\centering
\small
\setlength{\tabcolsep}{4pt}
\begin{tabularx}{\linewidth}{@{}>{\raggedright\arraybackslash}p{0.22\linewidth}>{\raggedright\arraybackslash}X>{\centering\arraybackslash}p{0.12\linewidth}>{\raggedright\arraybackslash}p{0.24\linewidth}@{}}
\toprule
Target & Task & Train/test & Score \\
\midrule
SpreadsheetBench & Edit Excel workbooks by generating and executing Python & 120/280 & Official cell- and sheet-level comparison \\
ALFWorld & Complete text-based household tasks through sequential actions & 3,553/134 & Success on official \texttt{valid\_unseen} games \\
OfficeQA & Retrieve Treasury Bulletin evidence and answer numerical questions & 50/172 & Normalized exact match \\
SearchQA & Answer questions from noisy retrieved web snippets & 400/1,400 & SQuAD-style exact match \\
LiveMath & Solve theorem-level five-choice mathematics questions & 35/125 & Exact choice accuracy \\
\bottomrule
\end{tabularx}
\caption{Benchmarks used: the train split is available during optimization, and the final target agent is evaluated on the test split.}
\label{tab:evaluation-targets}
\end{table}

\textbf{Metrics.} We report the optimizer, target, and finale costs, corresponding to the cost of the optimizer model, the target model during optimization, and the target model during evaluation on the full test split, respectively. Accuracy is the held-out accuracy on the benchmark. Table~\ref{tab:cost-comparison} combines the optimizer and target costs into one optimization cost; Table~\ref{tab:cost-breakdown} in the appendix gives the full breakdown.

\section{Results}\label{results}

\textbf{Overview.} Coding agents outperform the strongest published method on every benchmark at every combination of harness and budget (Table~\ref{tab:method-comparison}). While some benchmarks are already near-saturated with a high baseline score, we demonstrate significant improvement, with optimized agents obtaining near double the previous best on lower-scoring benchmarks such as LiveMath. The improvement over the best previous result ranges from 3\% on SearchQA to 98\% on LiveMath, averaging 0.15 in accuracy -- a 35\% improvement -- across the five benchmarks.

\begin{table}[!t]
\centering
\small
\renewcommand{\arraystretch}{1.10}
\setlength{\tabcolsep}{4pt}
\begin{tabularx}{\linewidth}{@{}l*{5}{>{\centering\arraybackslash}X}@{}}
\toprule
Method & Spreadsheet & OfficeQA & ALFWorld & SearchQA & LiveMath \\
\midrule
Baseline & 0.361 & 0.221 & 0.731 & 0.759 & 0.147 \\
Human skill & 0.429 & 0.459 & 0.567 & 0.772 & 0.288 \\
LLM skill & 0.368 & 0.366 & 0.657 & 0.761 & 0.280 \\
Trace2Skill & 0.407 & 0.209 & 0.828 & 0.786 & 0.328 \\
TextGrad & 0.382 & 0.300 & 0.709 & 0.775 & 0.272 \\
GEPA & 0.425 & 0.453 & 0.813 & 0.794 & 0.272 \\
SkillOpt & 0.475 & 0.488 & 0.858 & 0.802 & 0.328 \\
\midrule
Our baseline & 0.277 & 0.322 & 0.578 & 0.764 & 0.261 \\
\rowcolor{skillrow}Vanilla coding agents & \textbf{0.694} & 0.539 & \textbf{0.998} & \textbf{0.824} & 0.649 \\
\rowcolor{skillrow}\quad Two-phase & 0.636 & \textbf{0.546} & 0.944 & 0.823 & 0.595 \\
\rowcolor{skillrow}\quad with Agent Lightning skill & 0.667 & 0.545 & 0.947 & 0.819 & 0.656 \\
\rowcolor{skillrow}\quad with subagents & 0.525 & 0.515 & 0.971 & 0.817 & 0.572 \\
\rowcolor{skillrow}\quad with Git worktrees & 0.691 & 0.518 & 0.992 & 0.819 & \textbf{0.660} \\
\bottomrule
\end{tabularx}
\caption{Mean held-out accuracy by method and benchmark: all target agents use \texttt{gpt-5.4-mini}. The first seven rows are as reported by SkillOpt \citep{yang2026skillopt}. Our baseline is the unoptimized target agent. Optimized rows average the harness-budget means across all three harnesses and budgets; Subagents and Git worktrees average the \$10 runs of all three harnesses.}
\label{tab:method-comparison}
\end{table}

\textbf{Harnesses.} The difference in performance across harnesses can largely be attributed to per-run variance. No harness stands above the rest: Codex is ahead on SpreadsheetBench and LiveMath, Claude Code on OfficeQA, and the three are roughly even on ALFWorld and SearchQA (Appendix~\ref{appendix-detailed-results}). Codex and Copilot run the same model, \texttt{gpt-5.6-sol}, and the gap between them stays within the spread of repeated runs on every benchmark. The only benchmark with a consistent difference is LiveMath, where Claude Code, the only harness running \texttt{claude-opus-5}, scores lower than both; it is therefore more likely an effect of the model than of the harness in isolation.

\textbf{Skills.} The effect of giving the optimizer a skill is similarly trivial. Compared to the Vanilla version with the same benchmark, harness, and budget, the Agent Lightning skill scores lower more often than higher, and the same holds for two-phase optimization, Subagents, and Git worktrees (Table~\ref{tab:method-comparison}, Appendix~\ref{appendix-detailed-results}). Most of these differences, again, can be attributed to per-run variance, and no version consistently outperforms the rest. Guided optimizers do not find different solutions; they arrive at the same ones as the Vanilla version, slightly less often. For instance, on ALFWorld, nearly every Vanilla run removes the model from the final agent, whereas only about two thirds of two-phase runs and half of Subagents runs do so.

\begin{figure}[H]
\centering
\includegraphics[width=\linewidth]{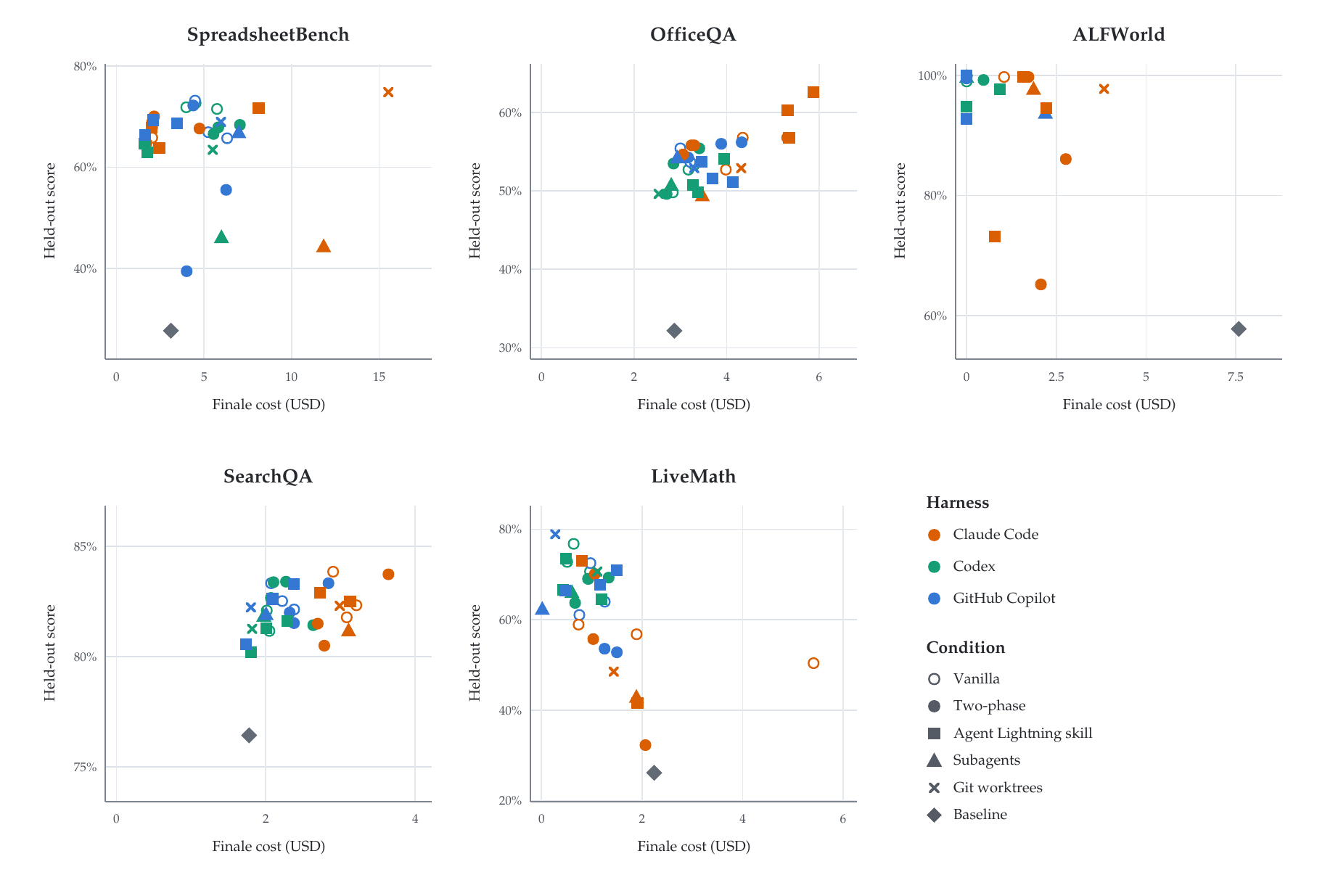}
\caption{Held-out score against finale cost: each point averages the runs of one harness and budget; the gray diamond is the unoptimized baseline.}
\label{fig:finale-cost-performance}
\end{figure}

\textbf{Cost and budget.} A larger budget does not directly translate to a better score. The per-run score is only weakly correlated with the amount spent: the \$25 runs beat the \$5 runs in just two thirds of the benchmark--harness pairs (Appendix~\ref{appendix-detailed-results}), and only by 0.03 on average. Similarly, runs are not always limited by their budget; at \$25, a quarter of the optimizers stop with more than a fifth of it unspent. A larger budget mainly gives the optimizer slack to recover from mistakes, and these are uncommon: on ALFWorld and LiveMath, the same solutions are found about as often at \$5 as at \$25. About two thirds of what is spent goes to the optimizer's own calls, and the rest to running the target on training examples (Table~\ref{tab:cost-breakdown}). Optimization spend is also largely independent of the finale cost (Table~\ref{tab:cost-comparison}, Figure~\ref{fig:finale-cost-performance}). On ALFWorld the test-time cost falls to virtually nothing, and on LiveMath it falls by about a third; on the other three benchmarks the optimized agent is more expensive than the baseline, despite the task prompt asking the optimizer to keep any increase in per-run cost minimal. Two-phase optimization, whose second phase exists to cut that cost, spends less on optimization but produces agents that are on average no cheaper to run.

\begin{table}[H]
\centering
\small
\renewcommand{\arraystretch}{1.12}
\setlength{\tabcolsep}{3pt}
\begin{tabularx}{\linewidth}{@{}lc*{5}{>{\centering\arraybackslash}X}@{}}
\toprule
Method & Budget & Spreadsheet & OfficeQA & ALFWorld & SearchQA & LiveMath \\
\midrule
Vanilla & \$5 & 6.46 / 3.94 & 5.34 / 3.33 & 5.34 / 0.35 & 5.07 / 2.55 & 4.91 / 2.38 \\
& \$10 & 9.76 / 4.70 & 9.95 / 3.51 & 9.74 / 0.00 & 8.96 / 2.39 & 9.32 / 1.22 \\
& \$25 & 22.96 / 3.51 & 23.52 / 3.94 & 23.27 / 0.52 & 20.32 / 2.40 & 19.96 / 0.78 \\
\addlinespace[3pt]
Two-phase & \$5 & 4.49 / 3.76 & 4.78 / 3.21 & 4.24 / 0.92 & 5.08 / 2.46 & 3.87 / 1.22 \\
& \$10 & 9.97 / 4.98 & 7.35 / 3.44 & 6.33 / 0.69 & 7.11 / 2.55 & 4.75 / 1.29 \\
& \$25 & 16.04 / 5.16 & 17.26 / 3.33 & 12.25 / 0.73 & 10.08 / 2.89 & 8.00 / 1.08 \\
\bottomrule
\end{tabularx}
\caption{Mean cost per run in USD: each cell reports optimization cost / finale cost, averaged over all three harnesses at the stated nominal budget.}
\label{tab:cost-comparison}
\end{table}

\textbf{Behaviour.} Optimizer behaviour is mostly consistent across runs. On SpreadsheetBench most runs add a repair step that retries after an execution failure; on OfficeQA they clean up the form of the answer and add further model calls; on SearchQA they rework how the retrieved snippets are presented and normalize the extracted answer. The edits are small, typically around two or three files and a couple of hundred lines, and are tested only a handful of times before the run ends.

ALFWorld and LiveMath are different, in that the gain comes from removing the model from the final solution. On ALFWorld almost every run replaces model-driven action selection with a deterministic controller, so that the final agent makes no model calls at all. On LiveMath most runs answer a subset of the test set from the answer selection structure alone and call the model only for the rest; these runs score well above the few that still query the model for every question. Such solutions raise the benchmark score without making the agent better in any general sense, a point we return to in the limitations.

\section{Limitations and future work}\label{limitations-and-further-work}

All five benchmarks provide labeled training examples, an automatic scorer, and a single scalar metric; this vastly simplifies the hill-climbing by giving the optimizer a clear relationship between its changes and the result. This is certainly not representative of many real-world scenarios with sparse, noisy, or abstract rewards. The same property allows benchmark-specific solutions: on ALFWorld and LiveMath, the gain comes from removing the model from the final agent, which improves the score without necessarily improving the agent on the task itself. And as Claude Code runs a different model from Codex and Copilot, the harness comparison also cannot be made in isolation from the model. In addition, all runs use a single target model with at most three repetitions per condition, and the published results we compare against are taken as reported rather than reproduced.

As for future work, while the Agent Lightning skill was developed over many iterations of working with Claude and Codex, the same setup could be applied to the skill itself with an optimizer recursively improving the skill. A full cross-model-harness evaluation would disentangle the effect of the harness from that of the model, and evaluating a target under a different harness from the one it was optimized in would show whether the improvements transfer. Lastly, each run in this work is independent, starting without context from previous runs and ending with a frozen target; an optimizer that retains what it has learned across runs and continues to refine the agent as the task environment shifts would constitute a step towards continual learning.

\section{Conclusion}\label{conclusion}

In this paper, we evaluate coding agents as optimizers of other agents. Given only a target agent, a train split, and a budget, a coding agent without guidance outperforms the strongest published skill-optimization method on all five benchmarks at every combination of harness and budget. The choice of harness, the addition of a skill, and the size of the budget all matter little: most of the differences between them can be attributed to per-run variance, and guided optimizers arrive at the same solutions as unguided ones. More broadly, this shows that agents are already capable of doing such work on their own, as long as they are given a metric to measure against.

{\small
\bibliographystyle{unsrtnat}
\bibliography{references}
}

\clearpage
\appendix
\section{Optimizer prompts}\label{appendix-prompts}

Every optimizer receives the task prompt below, shown for OfficeQA at the \$5 budget; only the package name, evaluation command, and dollar amount change across benchmarks and budgets.

{\footnotesize
\begin{verbatim}
I've got a coding agent in this workspace (/workspace) -- the `officeqa_agent` package
-- and it's underperforming on our benchmark. Can you raise its benchmark score while
keeping any increase in per-run cost minimal -- buy score cheaply, and only pay more
when it clearly earns its keep?

A few things to know:
- Measure your score on the labeled data by running:
    python -m officeqa_agent.run_baseline --dataset-root /data/train --model
    gpt-5.4-mini --artifacts-dir ./_outputs
- The labeled examples at `/data/train` are the only data present -- split your own
  validation subset out of them if you want one; the held-out scoring set is sealed and
  never here.
- Keep the agent's command-line interface working -- an external evaluator runs it the
  same way (`python -m officeqa_agent.run_baseline --dataset-root <inputs> --model
  gpt-5.4-mini --no-score`), so don't break that contract.
- Evaluation runs your agent with a fixed model, so changing the model in the agent's
  code won't move its score -- develop with anything you like, but the gains that count
  are the ones that lift accuracy at that fixed setting.
- Network egress is limited to the metered model gateway whose credentials are already
  in your environment.
- You have a budget of up to $5 for this work -- pace yourself so you finish within it.
  The dollar budget counts every API call this work makes -- both your own usage and the
  `officeqa_agent` agent's usage each time you evaluate it -- so factor re-evaluations
  into your pacing.
\end{verbatim}
}

Every other condition adds ``Use the agl-optimizer skill to guide how you approach this.'' to this prompt and provides a skill: the published one for the Agent Lightning skill condition (\url{https://github.com/microsoft/agent-lightning}), and the following for Two-phase, where the harness sends the phase switch once about half of the budget has been used:

{\footnotesize
\begin{verbatim}
Use the development budget in two phases:

1. Until the harness sends `[PHASE SWITCH]`, maximize benchmark score. Measure, iterate,
   and preserve the best-scoring checkpoint.
2. At the switch, treat that score as a constraint. Use the remaining budget to minimize
   the deployed agent's per-evaluation LLM cost, keeping only changes that preserve the
   best measured score.

Development spend and deployed-agent cost are different. Keep evaluating both score and
deployed cost, leave the best score-first/cost-second checkpoint on disk, preserve the
external interface, and never use held-out data or modify the scorer.
\end{verbatim}
}

Subagents and Git worktrees use variants of the Agent Lightning skill, each with one additional section. The section added for Subagents is:

{\footnotesize
\begin{verbatim}
Split the development budget across two subagents, reserving enough for final
selection. Run them concurrently as independent optimization threads from the
same starting checkpoint in isolated workspaces. Compare them using the same
evaluation setup. Select the better result, or combine only measured improvements
and re-evaluate the combined result before finalizing.

Use the harness's subagent tool directly (`Agent` in Claude Code, or its
equivalent). Choose each child model to fit its task when model selection is
available. Do not substitute background shell jobs or sequential main-thread
work for the two subagent calls. If the tool is unavailable, continue in the
main thread and state that limitation.
\end{verbatim}
}

The section added for Git worktrees is:

{\footnotesize
\begin{verbatim}
Use `/workspace` as the main checkout and `/worktrees` only for candidate checkouts.

1. In `/workspace`, initialize Git if needed, set the supplied neutral identity,
   stage the pristine target, and commit it before editing.
2. Create at least two candidate branches and attach them as worktrees under
   `/worktrees`. Start every candidate from the same pristine commit.
3. Give each candidate one focused hypothesis. Evaluate every candidate with the
   same training cases, command, target model, and effort unless effort itself is
   the tested change.
4. Record each candidate's commit, hypothesis, training score, evaluation cost,
   completion rate, and important failures in `/artifacts/worktree_candidates.md`.
5. Compare measured results rather than recency. Merge, cherry-pick, or restore the
   best candidate into `/workspace`, then re-run a final validation there.
6. Before exiting, remove every active candidate with `git worktree remove`, prune
   stale metadata, and leave `/workspace` at the selected measured checkpoint.

Do not use branches as unlabeled scratch space. Commit before scoring, do not compare
candidates evaluated under different conditions, and never spend held-out data.
\end{verbatim}
}

\section{Detailed results}\label{appendix-detailed-results}

\begin{figure}[!p]
\centering
\includegraphics[width=\linewidth]{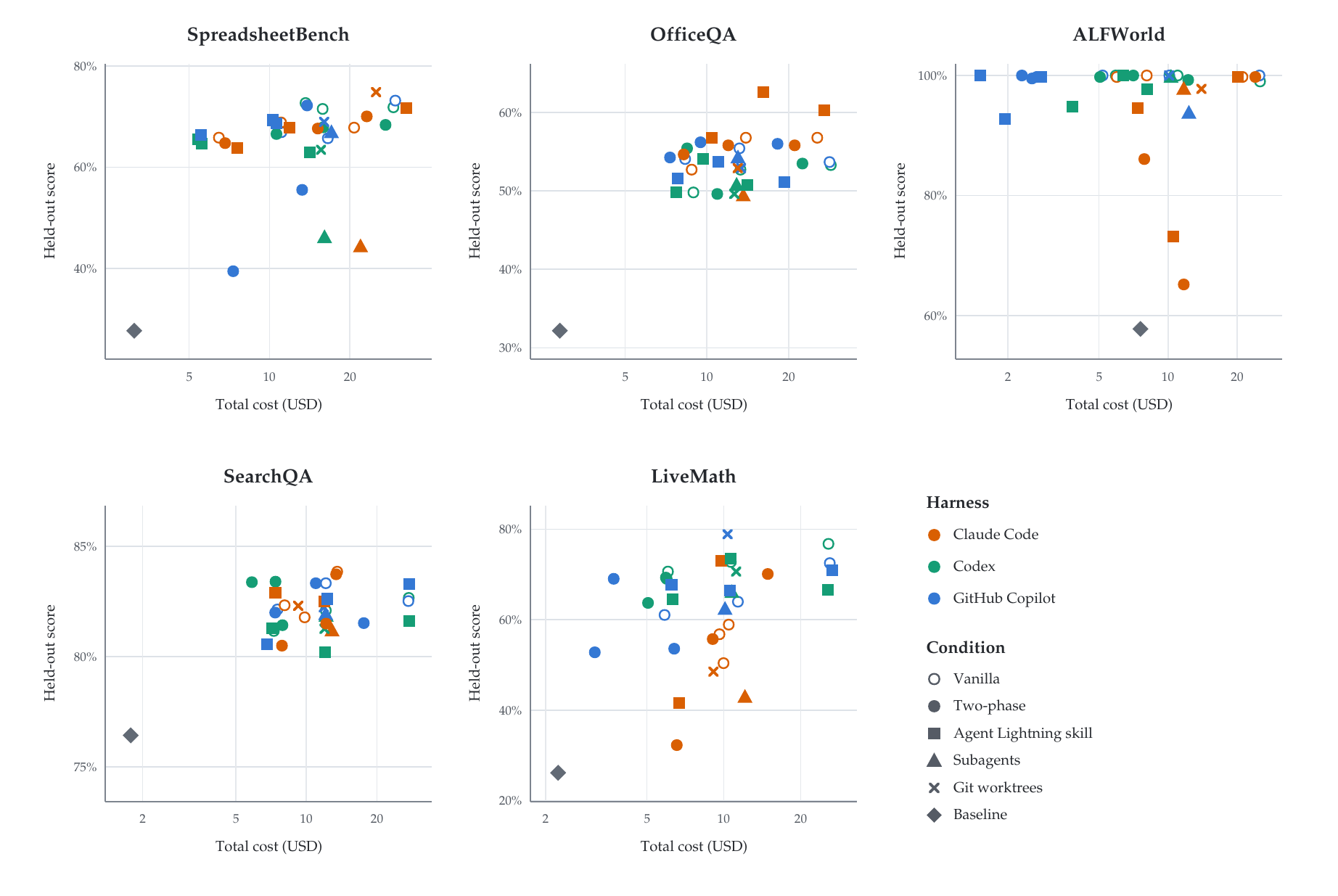}
\caption{Held-out score against total cost: total cost combines the optimization and finale costs, on a logarithmic axis.}
\label{fig:total-cost-performance}
\end{figure}

The following tables report mean held-out accuracy for each available benchmark--harness--budget condition. Each value averages up to three repetitions; dashes indicate unavailable results. Figure~\ref{fig:total-cost-performance} plots the held-out score against total cost.

\begin{table}[H]
\centering
\small
\renewcommand{\arraystretch}{1.12}
\setlength{\tabcolsep}{3pt}
\begin{tabular*}{\linewidth}{@{\extracolsep{\fill}}l*{9}{r}@{}}
\toprule
\multicolumn{10}{@{}l}{\textbf{Vanilla}} \\
& \multicolumn{3}{c}{Claude Code} & \multicolumn{3}{c}{Codex} & \multicolumn{3}{c}{Copilot} \\
Benchmark & \$5 & \$10 & \$25 & \$5 & \$10 & \$25 & \$5 & \$10 & \$25 \\
\midrule
SpreadsheetBench & 0.658 & 0.688 & 0.678 & 0.727 & 0.715 & 0.718 & 0.669 & 0.657 & 0.731 \\
OfficeQA & 0.527 & 0.568 & 0.568 & 0.498 & 0.527 & 0.533 & 0.541 & 0.554 & 0.537 \\
ALFWorld & 0.998 & 1.000 & 0.998 & 1.000 & 1.000 & 0.990 & 1.000 & 1.000 & 1.000 \\
SearchQA & 0.823 & 0.818 & 0.839 & 0.812 & 0.821 & 0.827 & 0.821 & 0.833 & 0.825 \\
LiveMath & 0.504 & 0.568 & 0.589 & 0.707 & 0.728 & 0.768 & 0.611 & 0.640 & 0.725 \\
\bottomrule
\end{tabular*}
\end{table}

\begin{table}[H]
\centering
\small
\renewcommand{\arraystretch}{1.12}
\setlength{\tabcolsep}{3pt}
\begin{tabular*}{\linewidth}{@{\extracolsep{\fill}}l*{9}{r}@{}}
\toprule
\multicolumn{10}{@{}l}{\textbf{Two-phase}} \\
& \multicolumn{3}{c}{Claude Code} & \multicolumn{3}{c}{Codex} & \multicolumn{3}{c}{Copilot} \\
Benchmark & \$5 & \$10 & \$25 & \$5 & \$10 & \$25 & \$5 & \$10 & \$25 \\
\midrule
SpreadsheetBench & 0.647 & 0.676 & 0.700 & 0.665 & 0.679 & 0.683 & 0.394 & 0.722 & 0.555 \\
OfficeQA & 0.547 & 0.558 & 0.558 & 0.554 & 0.496 & 0.535 & 0.543 & 0.562 & 0.560 \\
ALFWorld & 0.861 & 0.652 & 0.998 & 0.998 & 1.000 & 0.993 & 0.995 & 1.000 & 0.998 \\
SearchQA & 0.805 & 0.815 & 0.837 & 0.834 & 0.834 & 0.814 & 0.820 & 0.833 & 0.815 \\
LiveMath & 0.323 & 0.557 & 0.701 & 0.637 & 0.693 & 0.691 & 0.691 & 0.528 & 0.536 \\
\bottomrule
\end{tabular*}
\end{table}

\begin{table}[H]
\centering
\small
\renewcommand{\arraystretch}{1.12}
\setlength{\tabcolsep}{3pt}
\begin{tabular*}{\linewidth}{@{\extracolsep{\fill}}l*{9}{r}@{}}
\toprule
\multicolumn{10}{@{}l}{\textbf{Agent Lightning skill}} \\
& \multicolumn{3}{c}{Claude Code} & \multicolumn{3}{c}{Codex} & \multicolumn{3}{c}{Copilot} \\
Benchmark & \$5 & \$10 & \$25 & \$5 & \$10 & \$25 & \$5 & \$10 & \$25 \\
\midrule
SpreadsheetBench & 0.638 & 0.677 & 0.717 & 0.655 & 0.646 & 0.629 & 0.663 & 0.693 & 0.687 \\
OfficeQA & 0.568 & 0.626 & 0.603 & 0.498 & 0.541 & 0.508 & 0.516 & 0.537 & 0.512 \\
ALFWorld & 0.945 & 0.731 & 0.998 & 0.948 & 1.000 & 0.978 & 0.928 & 1.000 & 0.998 \\
SearchQA & 0.829 & 0.825 & -- & 0.813 & 0.802 & 0.816 & 0.806 & 0.826 & 0.833 \\
LiveMath & 0.416 & 0.731 & -- & 0.645 & 0.736 & 0.667 & 0.677 & 0.664 & 0.709 \\
\bottomrule
\end{tabular*}
\end{table}

\begin{table}[H]
\centering
\small
\renewcommand{\arraystretch}{1.12}
\setlength{\tabcolsep}{6pt}
\begin{tabular*}{\linewidth}{@{\extracolsep{\fill}}lrrr@{}}
\toprule
\multicolumn{4}{@{}l}{\textbf{Subagents (\$10)}} \\
Benchmark & Claude Code & Codex & Copilot \\
\midrule
SpreadsheetBench & 0.444 & 0.462 & 0.669 \\
OfficeQA & 0.494 & 0.508 & 0.543 \\
ALFWorld & 0.978 & 0.998 & 0.938 \\
SearchQA & 0.812 & 0.819 & 0.819 \\
LiveMath & 0.429 & 0.661 & 0.624 \\
\bottomrule
\end{tabular*}
\end{table}

\begin{table}[H]
\centering
\small
\renewcommand{\arraystretch}{1.12}
\setlength{\tabcolsep}{6pt}
\begin{tabular*}{\linewidth}{@{\extracolsep{\fill}}lrrr@{}}
\toprule
\multicolumn{4}{@{}l}{\textbf{Git worktrees (\$10)}} \\
Benchmark & Claude Code & Codex & Copilot \\
\midrule
SpreadsheetBench & 0.748 & 0.634 & 0.689 \\
OfficeQA & 0.529 & 0.496 & 0.529 \\
ALFWorld & 0.978 & 0.998 & 1.000 \\
SearchQA & 0.823 & 0.813 & 0.822 \\
LiveMath & 0.485 & 0.707 & 0.789 \\
\bottomrule
\end{tabular*}
\end{table}

\begin{table}[H]
\centering
\small
\renewcommand{\arraystretch}{1.12}
\setlength{\tabcolsep}{3pt}
\begin{tabular*}{\linewidth}{@{\extracolsep{\fill}}ll*{6}{r}@{}}
\toprule
& & \multicolumn{3}{c}{Vanilla} & \multicolumn{3}{c}{Two-phase} \\
\cmidrule(lr){3-5}\cmidrule(l){6-8}
Benchmark & Budget & Optimizer & Target & Finale & Optimizer & Target & Finale \\
\midrule
SpreadsheetBench & \$5 & 3.78 & 2.68 & 3.94 & 2.34 & 2.15 & 3.76 \\
 & \$10 & 5.80 & 3.96 & 4.70 & 5.27 & 4.70 & 4.98 \\
 & \$25 & 15.42 & 7.55 & 3.51 & 8.64 & 7.39 & 5.16 \\
\addlinespace[3pt]
OfficeQA & \$5 & 3.05 & 2.29 & 3.33 & 2.40 & 2.37 & 3.21 \\
 & \$10 & 6.48 & 3.47 & 3.51 & 3.83 & 3.52 & 3.44 \\
 & \$25 & 14.52 & 9.00 & 3.94 & 8.84 & 8.42 & 3.33 \\
\addlinespace[3pt]
ALFWorld & \$5 & 4.90 & 0.44 & 0.35 & 3.28 & 0.96 & 0.92 \\
 & \$10 & 9.11 & 0.64 & 0.00 & 4.21 & 2.13 & 0.69 \\
 & \$25 & 19.74 & 3.53 & 0.52 & 7.31 & 4.94 & 0.73 \\
\addlinespace[3pt]
SearchQA & \$5 & 2.77 & 2.30 & 2.55 & 2.17 & 2.91 & 2.46 \\
 & \$10 & 5.12 & 3.83 & 2.39 & 2.68 & 4.43 & 2.55 \\
 & \$25 & 12.89 & 7.43 & 2.40 & 4.05 & 6.03 & 2.89 \\
\addlinespace[3pt]
LiveMath & \$5 & 2.80 & 2.11 & 2.38 & 2.02 & 1.85 & 1.22 \\
 & \$10 & 5.43 & 3.89 & 1.22 & 2.33 & 2.42 & 1.29 \\
 & \$25 & 15.36 & 4.60 & 0.78 & 3.37 & 4.63 & 1.08 \\
\bottomrule
\end{tabular*}
\caption{Cost breakdown per run in USD: the optimizer and target costs add up to the optimization cost in Table~\ref{tab:cost-comparison}.}
\label{tab:cost-breakdown}
\end{table}

\end{document}